\documentclass[11pt, a4paper, onecolumn, logo, copyright]{googledeepmind}
\pdfoutput=1

\pdftrailerid{redacted}

\makeatletter
\renewcommand\bibentry[1]{\nocite{#1}{\frenchspacing\@nameuse{BR@r@#1\@extra@b@citeb}}}
\makeatother

\usepackage[authoryear, sort&compress, round]{natbib}
\usepackage{amsmath}

\title{Mixture-of-Depths: Dynamically allocating compute in transformer-based language models}

\correspondingauthor{draposo@google.com, adamsantoro@google.com}

\renewcommand{\today}{}

\author[1*]{David Raposo}
\author[1]{Sam Ritter}
\author[1,2]{Blake Richards}
\author[1]{Timothy Lillicrap}
\author[1]{Peter Conway Humphreys}
\author[1*]{Adam Santoro}

\affil[1]{Google DeepMind}
\affil[2]{McGill University \& Mila}
\affil[*]{Equal Contribution}

\begin{abstract}
Transformer-based language models spread FLOPs uniformly across input sequences. In this work we demonstrate that transformers can instead learn to dynamically allocate FLOPs (or \emph{compute}) to specific positions in a sequence, optimising the allocation along the sequence for different layers across the model depth. Our method enforces a total compute budget by capping the number of tokens ($k$) that can participate in the self-attention and MLP computations at a given layer. The tokens to be processed are determined by the network using a top-$k$ routing mechanism. Since $k$ is defined \emph{a priori}, this simple procedure uses a static computation graph with known tensor sizes, unlike other conditional computation techniques. Nevertheless, since the identities of the $k$ tokens are fluid, this method can expend FLOPs non-uniformly across the time and model depth dimensions. Thus, compute expenditure is entirely predictable in sum total, but dynamic and context-sensitive at the token-level. Not only do models trained in this way learn to dynamically allocate compute, they do so efficiently. These models match baseline performance for equivalent FLOPS and wall-clock times to train, but require a fraction of the FLOPs per forward pass, and can be upwards of 50\% faster to step during post-training sampling.
\end{abstract}

\begin{document}

\maketitle

\section{Introduction}
Not all problems require the same amount of time or effort to solve. Analogously, in language modeling not all tokens and sequences require the same time or effort to accurately make a prediction. And yet, transformer models expend the same amount of compute per token in a forward pass. Ideally, transformers would use smaller total compute budgets by not spending compute unnecessarily. 

Conditional computation is a technique that tries to reduce total compute by expending it only when needed \citep{bengio2013deep,bengio2013estimating,bengio2016conditional}. Various algorithms offer solutions to when and how much compute should be used \citep{ainslie2023colt5, fedus2022switch, bapna_controlling}. However, general formulations of this challenging problem may not work well with existing hardware constraints since they tend to introduce dynamic computation graphs \citep{graves_adaptive, dehghani2018universal}. The most promising conditional computation methods may instead be those that are harmonious with our current hardware stack, which prioritizes static computation graphs, and known tensor sizes that are selected to maximize hardware utilization.

Here we consider the problem of language modeling using a \textit{static} compute budget that can be made less than that used by a vanilla transformer. The network must learn how to \textit{dynamically allocate} the available compute by making decisions per-token, in each layer, about where to spend compute from the available budget. In our implementation total compute is user defined and unchanging prior to training, rather than being a function of the network's on-the-fly decisions. Thus, hardware efficiency gains---such as reduced memory footprint, or reduced FLOPs per forward pass---can be anticipated and exploited ahead of time. As we will show, these gains can be had without sacrificing overall performance.

\begin{figure}[h]
    \centering
    \includegraphics[width=0.95\textwidth]{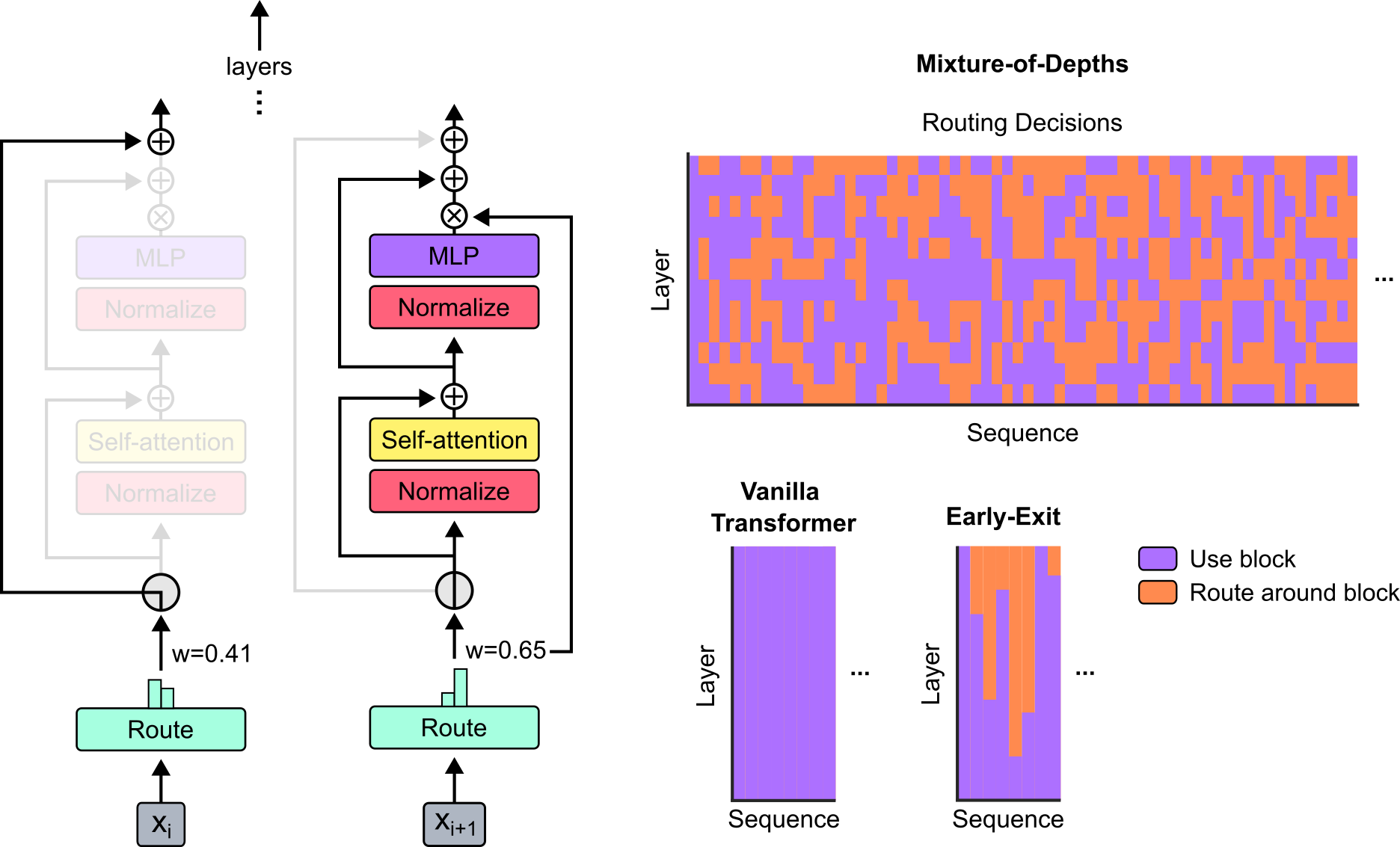}
    \caption{\textbf{Mixture-of-Depths Transformer.} As in mixture-of-experts (MoE) transformers we use a router to choose among potential computational paths. But unlike in MoE transformers the possible choices are a standard block's computation (i.e., self-attention and MLP) or a residual connection. Since some tokens take this second route, Mixture-of-Depths (MoD) transformers have a smaller total FLOP footprint compared to vanilla or MoE transformers. On the top right is depicted a trained model's routing decisions for a short sequence truncated to 64 tokens for visualization purposes. When examining the choices one can find tokens processed by later blocks' layers, despite passing through relatively few total blocks throughout the model's depth. This is a unique feature of MoD compared to conventional halting-based, or "early-exit" conditional computation, which instead engage blocks serially, or vanilla transformers, which engage every block.}
    \label{fig:mixture-of-depths}
\end{figure}

We leverage an approach akin to Mixture of Experts (MoE) transformers, in which dynamic token-level routing decisions are made across the network depth. Departing from MoE, we choose to either apply a computation to a token (as would be the case for a standard transformer), or pass it through a residual connection (remaining unchanged and saving compute). Also in contrast to MoE, we apply this routing to both forward MLPs and multi-head attention. Since this therefore also impacts the keys and queries we process, the routing makes decisions not only about which tokens to update, but also which tokens are made available to attend to. We refer to this strategy as Mixture-of-Depths (MoD) to emphasize how individual tokens pass through different numbers of layers, or blocks, through the depth of the transformer (see figure \ref{fig:mixture-of-depths}).

The MoD technique also allows one to trade-off performance with speed. On the one hand, one can train an MoD transformer that improves upon vanilla transformers by as much as $1.5\%$ on the final log probability training objective for equivalent training FLOPs (isoFLOP), and while taking an equivalent amount of wall-clock time to train. On the other hand, one can train an MoD transformer that achieves training loss parity with an isoFLOP optimal vanilla transformer, but which uses a fraction of the FLOPs (upwards of 50\%) per forward pass, and hence is faster to step. Together, these results imply that MoD transformers learn to route intelligently (i.e., skipping computations that are unnecessary) since they can achieve equal or better log probabilities per sequence despite a smaller FLOP footprint per forward pass. 

\section{Background}
The transformer architecture has become the workhorse of a revolution in practical artificial intelligence, bringing unprecedented capabilities at the cost of expensive training runs and serving procedures. This has spurred tremendous interest in making transformer architectures more efficient \citep{tay_efficient, gupta2021compression}. One of the promising approaches is \textit{conditional computation}, whereby learned mechanisms determine when and how to expend computation. This terminology was introduced by \citet{bengio2013deep}, and the concept was explored further over the next several years \citep{bengio2013estimating, cho2014exponentially, graves_adaptive, jernite2017variable, bengio2016conditional, wang_skipnet}.

A wide variety of recent work has developed conditional computation methods for transformers. Some of this work focuses on "early exiting", that is, learning to decide when to end computation on a given token, allowing the token to skip any remaining transformer layers after the exit decision is made \citep{elbayad_depth, liu2021anytime, schuster2022confident}. In MoD, unlike in early-exit methods, a token can skip middle layers, then be updated via self-attention with tokens that that have gone through all the middle layers. We speculate that this might be a useful property.

Other work has developed methods for iterating transformer layers with shared weights for an adaptive number of steps \citep{simoulin-crabbe-2021-many, dehghani2018universal}. \citet{bolya2023token} developed a method for choosing tokens to merge when running inference on a trained vision transformer which notably requires no learning. \citet{lei2023conditional} make use of conditional computation in a fine tuning setting by building on adapter approaches \citep{he2021towards} to learn to skip blocks of frozen pre-trained weights in favor of running only a small fine-tuned adapter. 

CoLT5 \citep{ainslie2023colt5} uses conditional routing to select whether a given token will pass through a heavy or light pathway for each feedforward layer. Further, they use the same routing mechanism to select whether a token will attend to all other tokens or to a select few, as in \citet{guo2022longt5}. Like MoD, CoLT5 uses soft top-k for making routing decisions. However, CoLT5 focuses on a encoder-decoder setting, and thus does need to contend with the problem of efficient sequential decoding given the non-causal nature of the top-k operation. In contrast, our current work with MoD focuses on the decoder-only setting, and so we propose a predictive router to enable efficient inference for conditional computation in transformers.

One successful formulation of conditional computation is the the "mixture-of-experts" layer (MoE) as introduced by \citet{shazeer2017outrageously}. Developed initially in the context of LSTMs, later work showed compelling empirical results for MoE with transformers \citep{lepikhin2020gshard, fedus2022switch, zoph2022stmoe}. Unlike other conditional computation approaches that try to conserve or expend additional compute, MoE transformers use conditional logic to route tokens to one of many expert MLPs while keeping total compute expenditure constant. Our mixture-of-depths method can be thought of as using the routing logic from MoE transformers, but rather than having multiple experts, MoD deploys a single expert which can be dynamically skipped. 

\section{Implementing Mixture-of-Depths Transformers}
Our high-level strategy is as follows:
\begin{itemize}
    \item Set a static compute budget that is less than that of an equivalent vanilla transformer by limiting the number of tokens in a sequence that can participate in a block's computations (i.e., self-attention and subsequent MLP). For example, while a vanilla transformer might permit all the tokens in a sequence to participate in self-attention, we might limit the number to 50\% of the tokens in a sequence. See section \ref{sec:define-compute-budget}.
    \item Use a per-block router to emit a scalar weight for each token, which expresses the router's preference for that token to participate in a block's computations or to route around it. See section \ref{sec:routing-to-nowhere}.
    \item Identify the top-$k$ scalar weights (per sequence, per block) to select those tokens that will participate in a block's computations. Since precisely $k$ tokens will participate in the block's computations, the computation graph and tensor sizes remain static throughout training; it is merely the tokens' participation that is dynamic and context-sensitive, as determined by the router. See section \ref{sec:routing-scheme}.
\end{itemize}

We then discuss some complications when sampling post-training in section \ref{sec:sampling}.

\subsection{Defining a compute budget}
\label{sec:define-compute-budget}

To enforce a total compute budget per forward pass we leverage the notion of \textbf{capacity}, which defines the total number of tokens that comprise the input to a given computation (e.g., the tokens participating in self-attention, a given expert in MoE transformers, etc). For example, the self-attention and MLP in each vanilla transformer block have a capacity of $T$---the total number of tokens across the sequence and batch. MoE transformers, on the other hand, use a capacity less than $T$ per expert MLP so as to more evenly divide the total compute across each expert. But, since they use multiple experts per block, their total capacity is approximately equal to that of a vanilla transformer.

Generally, it is the token capacity that determines the total FLOPs for transformers that use conditional computation, rather than the outcomes of any routing decisions. This is because static-graph implementations account for the worst-case scenarios decisions; e.g., a computation's inputs will be padded to its capacity amount even if relatively few tokens actually end up routing to it, and/or tokens will be dropped from the computation if the capacity is exceeded.

We can achieve our goal of using a smaller compute budget per forward pass compared to a vanilla transformer by lowering the capacity of the computations. However, using a smaller compute budget haphazardly will result in a performance degradation. We hypothesize that \textit{certain} tokens might not require as much processing as others, and these tokens can be identified through learning. Therefore, if the network learns to choose the right tokens to fill up its capacities, then it may preserve its performance. In the following we describe routing schemes that can be used for this purpose.

\subsection{Routing around transformer blocks}
\label{sec:routing-to-nowhere}
We consider the setting whereby we route tokens to one of two computational paths: (1) self-attention and MLP blocks, and (2) a residual connection. The latter is computationally cheap, and results in a block output that is entirely determined by the value of its input. The former path is computationally expensive.

The total number of FLOPs per forward pass will be fewer than that in a vanilla transformer if we set the capacity for path (1) to be anything less than $T$ (the total number of tokens across the sequence and batch). For example, if we were to set a block's capacity to  $\frac{T}{2}$ (i.e., half the number of tokens as would be the case in a vanilla transformer) then query-times-key matrix multiplication during self-attention becomes $25\%$ as FLOP-intensive as in a vanilla transformer ($(\frac{T}{2})^2$ vs. $T^2$). Similar calculations can determine the FLOP-savings for the MLP.

Intuitively, the total FLOPs per forward pass decreases (and the time to complete a forward pass decreases) in proportion to how aggressively we shrink the blocks' capacities. However, downstream performance will also be affected by how aggressively we shrink the blocks capacities, and by the routing algorithm we implement. 

At one extreme, if we leave each block's capacity at $T$ and route every token to (rather than \emph{around}) each block, then we recover a vanilla transformer. At the other extreme, if we set each block's capacity to $0$ and route all tokens \emph{around} each block, then we're left with a very fast model that doesn't engage with the vast majority of the transformer's parameters, and undoubtedly has poor downstream performance. We hypothesize that somewhere between these two extremes is an optimal model that is faster than a vanilla Transformer and performs as well, if not better, all while being faster to step.

\subsection{Routing schemes}
\label{sec:routing-scheme}
Naively, one can leverage stochasticity to route tokens, akin to layer or block ``dropout''. We present this routing scheme as a control, and will show that it significantly under-performs relative to vanilla transformers.

We hypothesize that \emph{learned} routing is preferable. Intuitively, the network should be able to learn which tokens require more or less processing than others. If we are correct that Transformers often expend more compute than they need to make their predictions, then it is an empirical question as to how aggressively we can shrink each block's capacity, and hence, how many tokens we can afford to route around each block. 

\begin{figure}[h]
    \centering
    \includegraphics{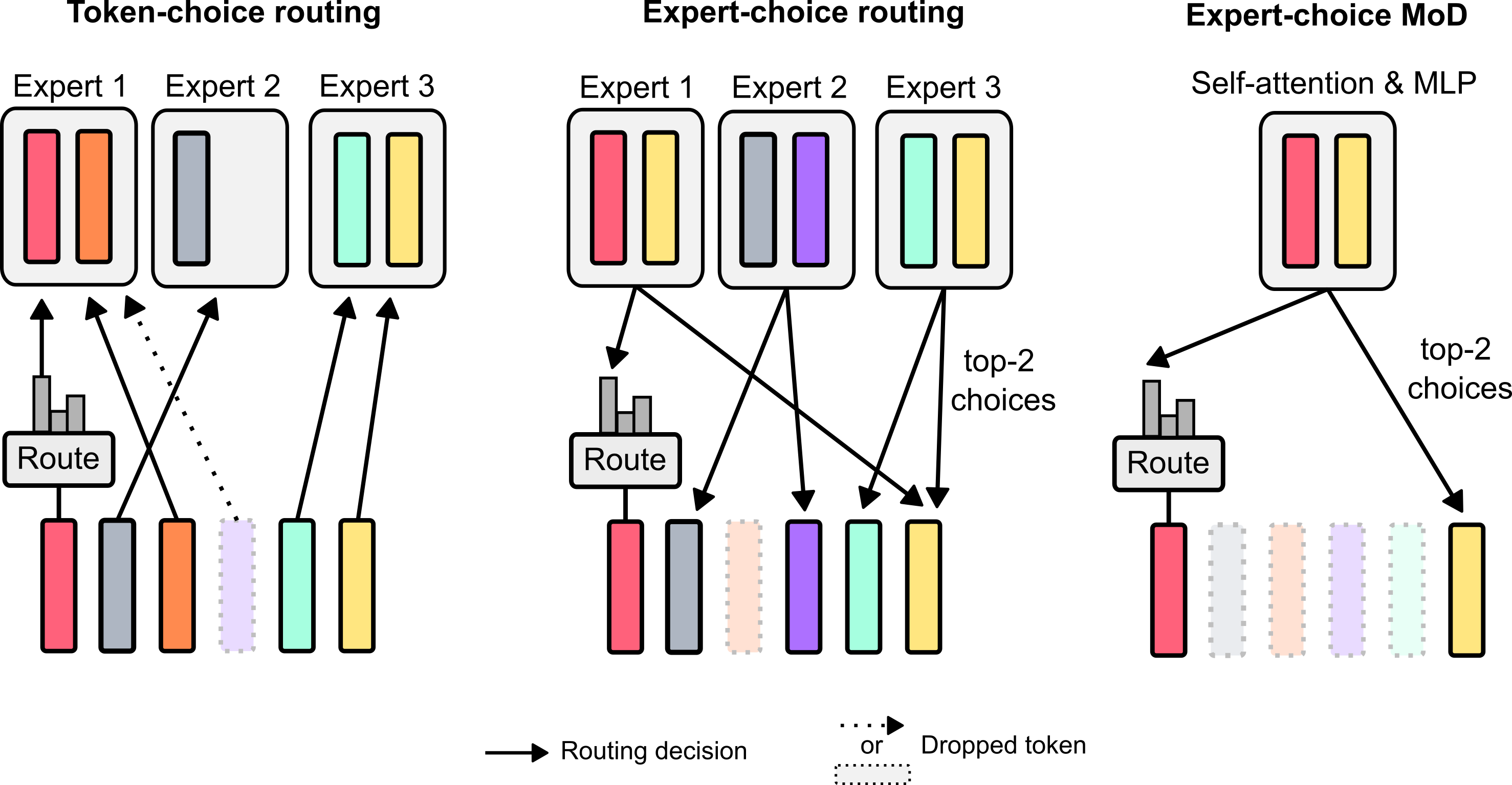}
    \caption{\textbf{Routing schemes.} Tokens are funnelled to the computational path of their choice when using token-choice routing (left). If a given path exceeds its capacity (e.g., more than two tokens in this example) then surplus tokens must be dropped (purple token). The exact token that is ultimately dropped depends on the  precise implementation in the underlying code. For example, priority is often given to those tokens that come earlier in the sequence or batch order. With expert-choice routing (middle), precisely $k$ (in this case, two) tokens are chosen per path using a top-$k$ mechanism across the tokens' router weights. Here, tokens are dropped if they are not among the top-$k$ with respect to any given path (orange token), and some tokens may even be funnelled to multiple paths (yellow token). In this work we deploy expert-choice routing (right). However, because we use just a single path, we \emph{leverage} the implicit knowledge that tokens will be dropped if $k$ is less than the sequence length so that we can route tokens away from the self-attention and MLP computations, thus expending fewer FLOPs in a given forward pass of the model.}
    \label{fig:routing}
\end{figure}

There are two learned routing schemes we consider (see figure \ref{fig:routing}): token-choice and expert-choice. In token-choice routing, a router produces per-token probability distributions across computational paths (e.g., across expert identities in MoE Transformers). Tokens are then shuttled to the path they prefer---i.e., that with the highest probability---and auxiliary losses ensure that all tokens don't converge to the same path. Token-choice routing can have load balancing problems since there isn't a guarantee that tokens divide themselves appropriately between the possible paths. ``Expert choice routing'' flips this recipe on its head: rather than having tokens choose the path they prefer, each path instead chooses the top-$k$ tokens based on the tokens' preferences. This ensures a perfect load balance since $k$ tokens are guaranteed to be shuttled to each path. However, it could result in over- or under-processing of some tokens, since some tokens may be among the top-$k$ for multiple paths, or for none of them.

We decided to leverage expert-choice routing for a few reasons. First, it obviates the need for an auxiliary balancing loss. Second, since the top-$k$ operation depends on the magnitude of the router weights, this routing scheme allows for relative routing weights to help determine which tokens most need the block's computations; routers can try to \emph{ensure} that the most critical tokens are among the top-$k$ by setting their weight appropriately, which is not possible with token-choice routing schemes. For our specific use-case, wherein one computational path is essentially a null operation, it might be critical that important tokens are routed away from the null operation. Third, because we only route through two paths, a single top-$k$ operation can efficiently split the tokens into two mutually exclusive sets, one for each computational path, preventing the over- or under-processing problem mentioned above. 

\subsection{Routing implementation}
\label{sec:routing-implementation}
As a reminder of the high-level intuition, each token is processed by a router to produce a scalar weight, and the top-$k$ weights are then used to choose the token identities that will route through a transformer's block, which comprises self-attention and the subsequent MLP. 

Suppose we have the set of token embeddings in a sequence of length $S$ for a given layer $l$; that is $X^l = \{x_i^l | i \text{ is an integer, }1 \leq i \leq S \}$.  The router weight for a given token embedding is a scalar produced as a result of a linear projection, $r_i^l=w_\theta^Tx_i^l$. 

Our goal is to use these router weights to determine the output of a block's computation of each token. Suppose $P_\beta(R^l)$ is the $\beta$-th percentile of the set of router weights $R^l$, where $\beta=1 - C/S$ and $C$ is the user-defined capacity per batch element (an integer $<S$ that defines the number of tokens from a sequence that will be processed by a given function). A block's output for a given token is:

\begin{align}
    x_i^{l+1} = 
    \begin{cases}
    r_i^l f_i(\tilde{X}^l) + x_i^l, & \text{if } r_i^l > P_\beta(R^l) \\
    x_i^l, & \text{if } r_i^l < P_\beta(R^l)  \\
    \end{cases}
\end{align}

Here, $\tilde{X}^l$ is the set of tokens whose router values $r_i^l > P_\beta(R^l)$ (that is, the ``top-k'' tokens), and $f$ comprises self-attention and the subsequent MLP. Note that the output for a given token $x_i^{l+1}$ might depend on other tokens $x_{i \neq j}^l$ because of the self-attention operation. The cardinality of $\tilde{X}^l$ is $C$ (or $k$): the user-defined capacity. Therefore, the mixture-of-depths transformer accrues compute savings relative to the baseline because the input to the block's computations $f$ comprise fewer tokens than usual ($C < S$), rendering the self-attention and MLP less expensive. 

Notably, we multiply the output of the function $f$ by the router weights. This puts the router weights along the ``gradient path'', thus subjecting them to the forces of gradient descent through the course of the language modeling task (We experimented with versions where the router weights are also included along the computational path for those tokens that bypass the block's computations, but it seems to be sufficient---and implementationally simpler---to only include the router weights along the computational path for those tokens that do not bypass the block's computations). 

\subsection{Sampling}
\label{sec:sampling}
While expert-choice routing has a number of advantages, it has one distinct problem: the top-$k$ operation is non-causal. This means that whether a given token's routing weight is among the top-$k$ for the sequence depends on the values of the routing weights for tokens that come after it, which we don't have access to when autoregressively sampling. 

We tested two methods to work around this problem. The first introduces a simple auxiliary loss that empirically affects the primary language modeling objective by approximately $0.2-0.3\%$, but allows us to sample from the model autoregressively. We  use a binary cross-entropy loss wherein the router's outputs provide the logits, and the top-$k$ selections of these logits provide the targets (i.e. 1 if a token was among the top-$k$, and 0 if not). Intuitively, this loss centers the sigmoid of the router's outputs around $0.5$; those tokens that are selected among the top-k are pressured to produce router outputs above $0.5$, and those not among the top-k will be pressured to produce router outputs below $0.5$. The second method introduces a small auxiliary MLP predictor (akin to a second router) that receives the same inputs as the router (with a stop gradient), but whose output is a prediction whether that token will be among the top-$k$ or not in the sequence. This method does not affect the language modeling objective, and empirically does not significantly impact the step speed.

Equipped with these new methods, we can sample autoregressively by choosing to route tokens \textit{to} or \textit{around} a block based on the router's output, which does not depend on any information from future tokens. We provide empirical evidence that this is a relatively easy auxiliary task that quickly achieves $99\%$ accuracy.

\subsection{Training methods}
All models use the same basic hyperparameter configurations (e.g. cosine schedules equal to $1 \times$ the training steps, 128 batch size, 2048 sequence length) except for changes to the number of layers, heads, and embedding size to produce differently sized models during isoFLOP analyses.  

\section{Results}
\subsection{Training, isoFLOP comparisons}
\begin{figure}[h]
    \centering
    \includegraphics[width=\textwidth]{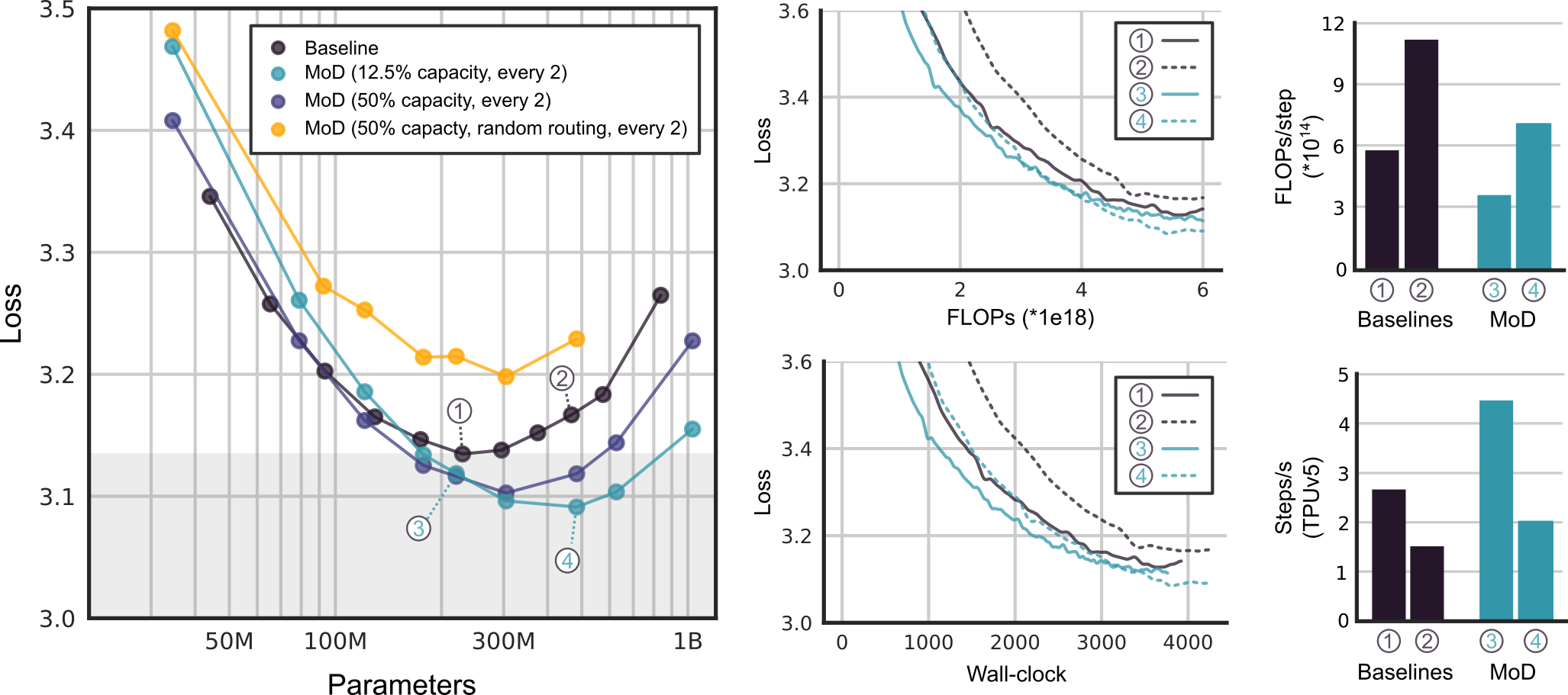}
    \caption{\textbf{MoD hyperparameter tuning.} Variants of the MoD transformer were trained for 6e18 FLOPs to determine the optimal hyperparameters for further isoFLOP analyses. On the left plot, the grey box indicates models that perform better than the isoFLOP optimal baseline. We found the best MoD variant to be that which has the option to route every other block, and which uses a top-k of $256$ (so, $256$, or 12.5\% of the sequence's tokens are processed by self-attention and the subsequent MLP, while $1792$ tokens, or $87.5\%$ of the sequence's tokens route \emph{around} the block). Shown on the right are the learning curves for a selection of models. Notably, model \#3 achieves equal performance to the isoFLOP optimal baseline but steps 66\% faster, due to the relatively fewer FLOPs needed per forward pass.}
    \label{fig:mod-learning-curve}
\end{figure}

We first trained models with a relatively small FLOP budget (6e18) to determine optimal hyperparameters (see figure  \ref{fig:mod-learning-curve}). In general, we found that MoD transformers drag the baseline isoFLOP curve "down and to the right". That is, the optimal MoD transformer achieves a lower loss than the optimal baseline, and also has more parameters. A fortunate consequence of this effect is that there exist smaller MoD models that, while they are not themselves isoFLOP optimal for their hyperparameter setting, are nevertheless as- or better-performing than the optimal baseline model while being faster to step. For example, a 220M parameter MoD (figure \ref{fig:mod-learning-curve} model \#3) variant slightly outperforms the isoFLOP optimal baseline (also 220M, figure \ref{fig:mod-learning-curve} model \#1), but is upwards of 60\% faster to step during training. Crucially, when run on equivalent hardware these two model variants take take approximately the same amount of wall-clock time to train (figure \ref{fig:mod-learning-curve}). 

We tested routing every block or every other block, using capacities from 12.5\% to 95\% of the total sequence. While routing every \textit{other} block was crucial for strong performance, we found that aggressive capacity reduction was best (gradual improvements were observed when reducing the capacity down to 12.5\% of the total sequence, corresponding to 87.5\% of tokens routing \textit{around} blocks, with performance degrading beyond this point). So, it seems the network is robust to significant capacity reductions as long as there is frequent opportunity for full capacity self-attention and MLP computations.

Learned routing is crucial, as MoD transformers that use stochastic routing (implemented using a top-$k$ operation on router weights sampled from a Gaussian distribution) perform drastically worse than both the baseline and normal MoD transformer (figure \ref{fig:mod-learning-curve}). 

\begin{figure}[h]
    \centering
    \includegraphics[width=\textwidth]{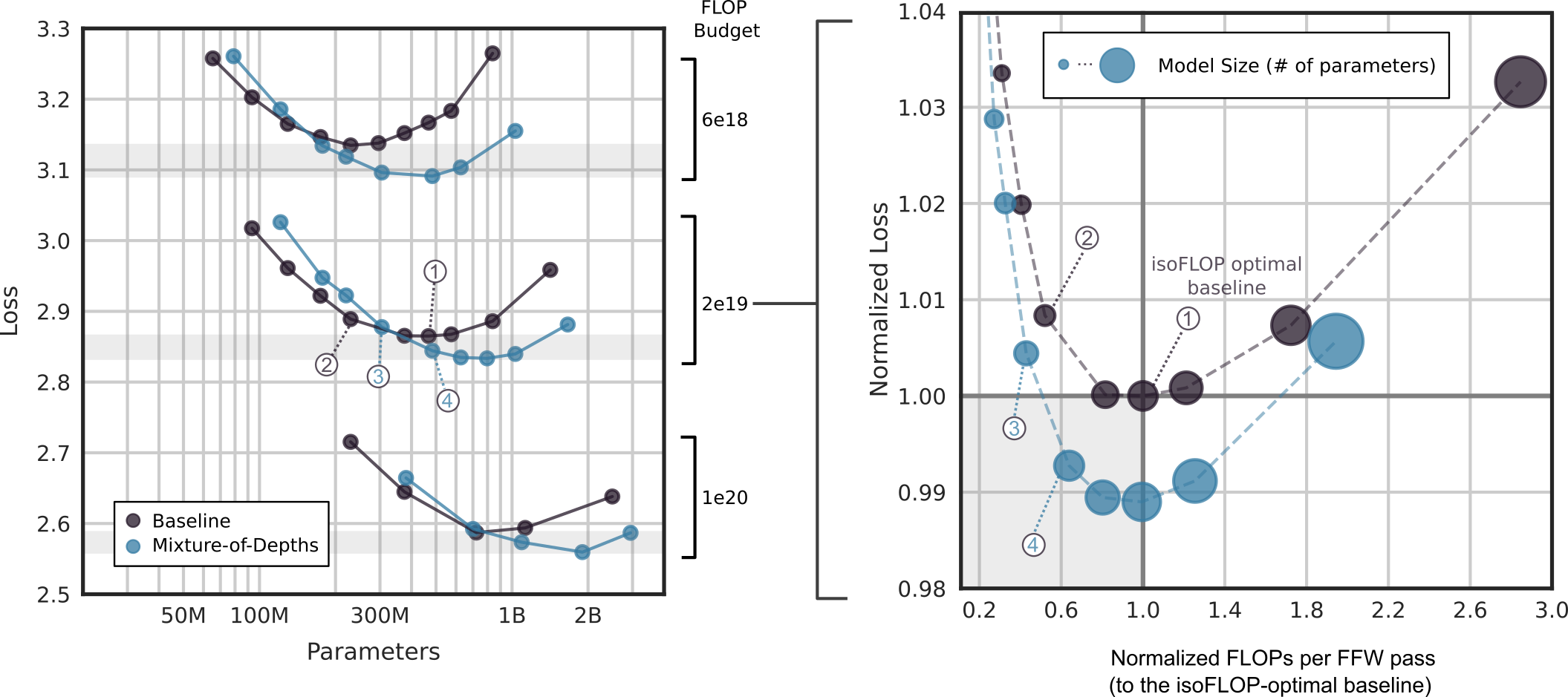}
    \caption{\textbf{isoFLOP analysis.} We used the 12.5\% capacity MoD variant to perform an isoFLOP analysis for 6e18, 2e19, and 1e20 FLOPs, training models varying in size from 60M to 3B parameters. Depicted on the right are the relative FLOPs per forward pass (normalized to the isoFLOP optimal baseline). There exist MoD variants that are both faster to step (by virtue of requiring fewer FLOPs per forward pass) and better performing than the isoFLOP optimal baseline.}
    \label{fig:isoflop}
\end{figure}

Depicted in figure \ref{fig:isoflop} is an isoFLOP analysis for 6e18, 2e19, and 1e20 total FLOPs. The trend that FLOP-optimal MoD transformers have more parameters than the baseline continues for these larger FLOP budgets. Notably, there exist MoD variants that are appreciably faster to step than the isoFLOP-optimal baseline (measured as steps per second when training on equivalent hardware) while also achieving a lower loss (in figure \ref{fig:isoflop} we depict normalized FLOPs per forward pass rather than wall-clock step time \textit{per se}, but from our experiments the two are tightly correlated. A similar plot can be produced showing relative wall-clock step times and the same basic trend is present). 

Step-wise speed gains come from two sources. First, the FLOP-per-parameter ratio in MoD transformers is less than in the baselines because some proportion of tokens are routed around blocks. So, for a given model size, a transformer requires fewer FLOPs per forward pass. Second, since isoFLOP-optimal MoD transformers are both bigger and achieve a lower loss than the isoFLOP-optimal baseline, there exist smaller MoD variants that perform as well or better than the isoFLOP-optimal baseline, and these variants are faster to step because they are smaller. Altogether, then, there exist MoD transformers that perform as well as isoFLOP-optimal baselines and are faster to step, both because they use fewer FLOPs per parameter and because they use fewer parameters.

Figure \ref{fig:isoflop} also reveals another important finding: the optimal MoD transformer is that which uses as many FLOPs per forward pass as the isoFLOP optimal baseline. This finding allows one to directly predict which sized MoD transformer will perform optimally for a given isoFLOP training budget: one just needs to tune the model size for a given MoD configuration (i.e., capacity and routing frequency) to produce a model that uses as many FLOPs per forward pass as the isoFLOP-optimal baseline, and they will have the optimally performing MoD variant for that configuration. Empirically, we find that it is better to add depth than to add width when adding FLOPs to the model.

Nevertheless, while the FLOPs per forward pass determines which model will be the isoFLOP optimal, it does not predict whether the optimal loss will improve upon the baseline (see figure \ref{fig:mod-learning-curve}. Namely, the optimal capacity appears to be empirically determinable. We found that it is best to use 12.5\% capacity blocks, every other block. 

We noticed that MoD transformers had memory savings relative to equivalently sized baseline models at larger sizes, with some variants requiring fewer total devices (i.e., a smaller TPU topology). We did not study this extensively, but we anticipate that as one scales to larger models, these savings could be an important consideration when choosing model variants to train, and could have significant positive effects in regards to the KV cache size during autoregressive sampling. 

\begin{figure}[h]
    \centering
    \includegraphics[width=\textwidth]{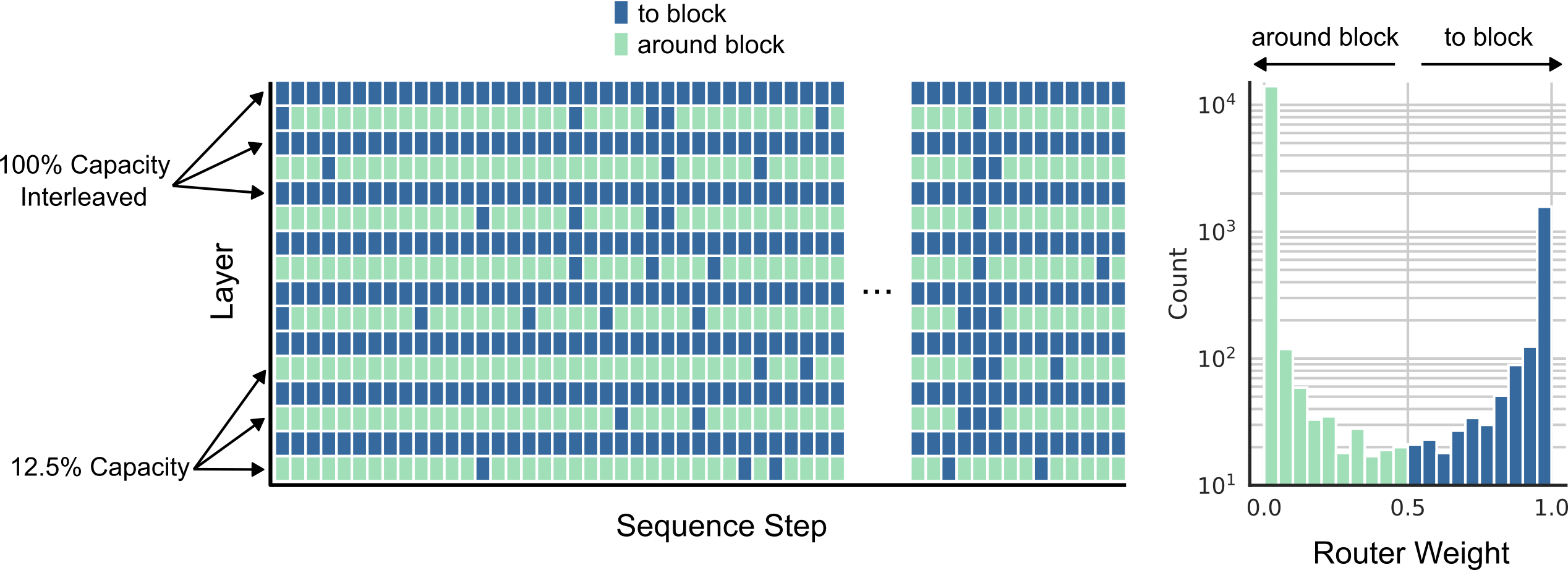}
    \caption{\textbf{Routing analysis.} We trained an MoD transformer that interleaved $12.5\%$ capacity routing blocks with full-attention blocks. As expected, the number of tokens that route to (rather than around) a block is sparse in routing blocks, though the network does sometimes preferentially route certain tokens to each block along its depth. This can be seen in the left figure that depicts routing decisions, where we observe a vertical band of dark blue towards the end of the sequence. As expected, the distribution of router weights are as the auxiliary loss dictates: approximately $12.5\%$ of weights are above 0.5 and $87.5\%$ are below (histogram, right).}
    \label{fig:routing-analysis}
\end{figure}

Figure \ref{fig:routing-analysis} shows the routing decisions for an MoD transformer trained with interleaved routing blocks. Despite aggressive routing around the blocks, transformers are able to achieve performance improvements relative to baselines. We observe patterns that might warrant further study; namely, some tokens appear to engage each block along the transformer's depth, while others decide to route around blocks whenever possible. Preliminary analyses suggest that the tokens that engage with blocks more frequently are correlated with output predictions that have higher entropy, which possibly corresponds to predictions that are more difficult to make.

\subsection{Auto-regressive Evaluation}
\begin{figure}[h]
    \centering
    \includegraphics[width=\textwidth]{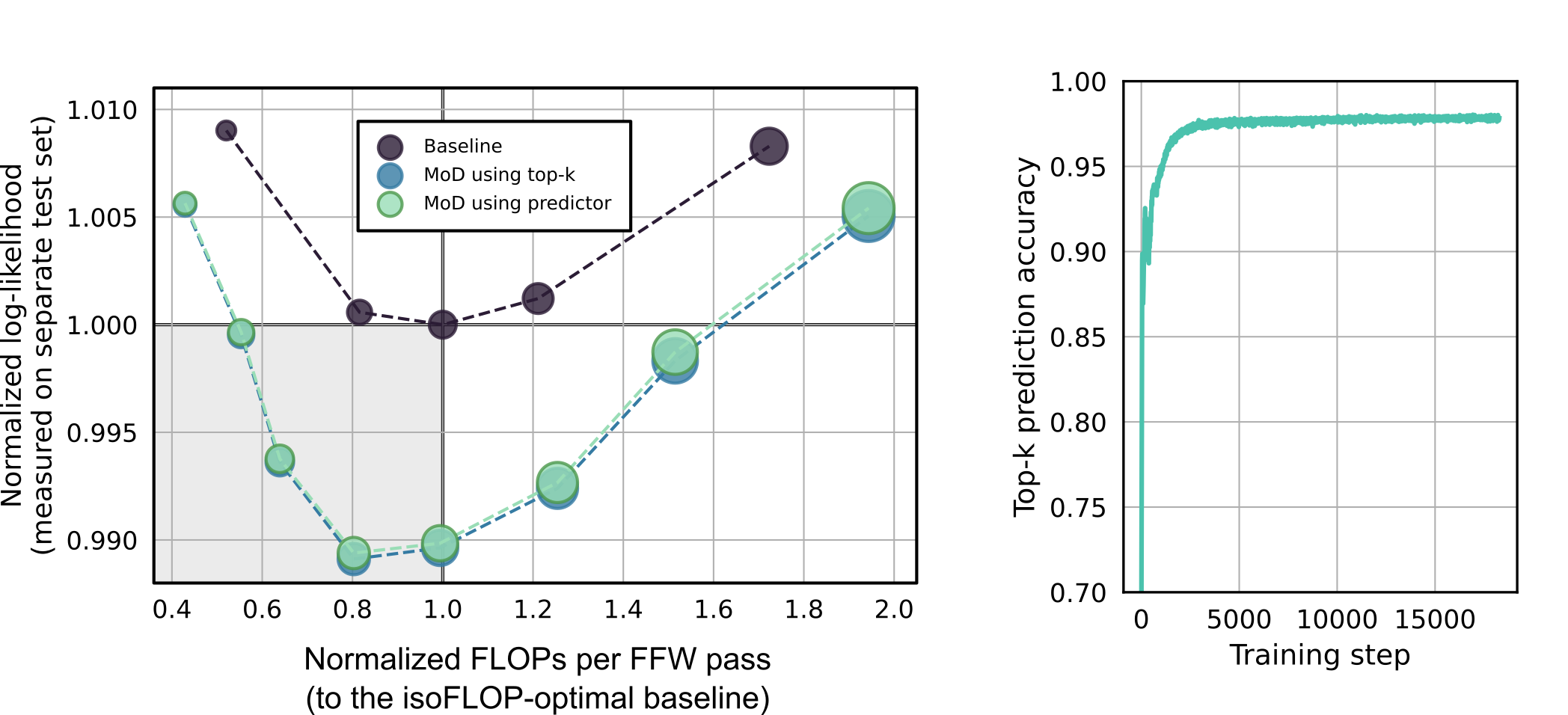}
    \caption{\textbf{Auto-regressive evaluation.} Switching from the non-causal top-$k$ routing scheme in training to a causal predictor-based approach during auto-regressive sampling leads to minimal performance degradation. This is perhaps due to the ease of learning this prediction problem, which is upwards of 97\% accurate soon into training.}
    \label{fig:autoregressive}
\end{figure}

We evaluated MoD variants during auto-regressive sampling (see figure \ref{fig:autoregressive}). Each model was tested on exactly the same held-out data comprising $256000$ sequences ($500$M tokens). When switching from the top-$k$ routing method to the predictor-based routing method we observed little performance degradation. As in the training setting, there exist MoD variants that are better performing than the isoFLOP-optimal baseline, while requiring fewer FLOPs per forward pass. These results suggest that the compute savings offered by MoD transformers should translate beyond the training setting.

\subsection{Mixture-of-Depths-and-Experts (MoDE)}
\begin{figure}[h]
    \centering
    \includegraphics[width=\textwidth]{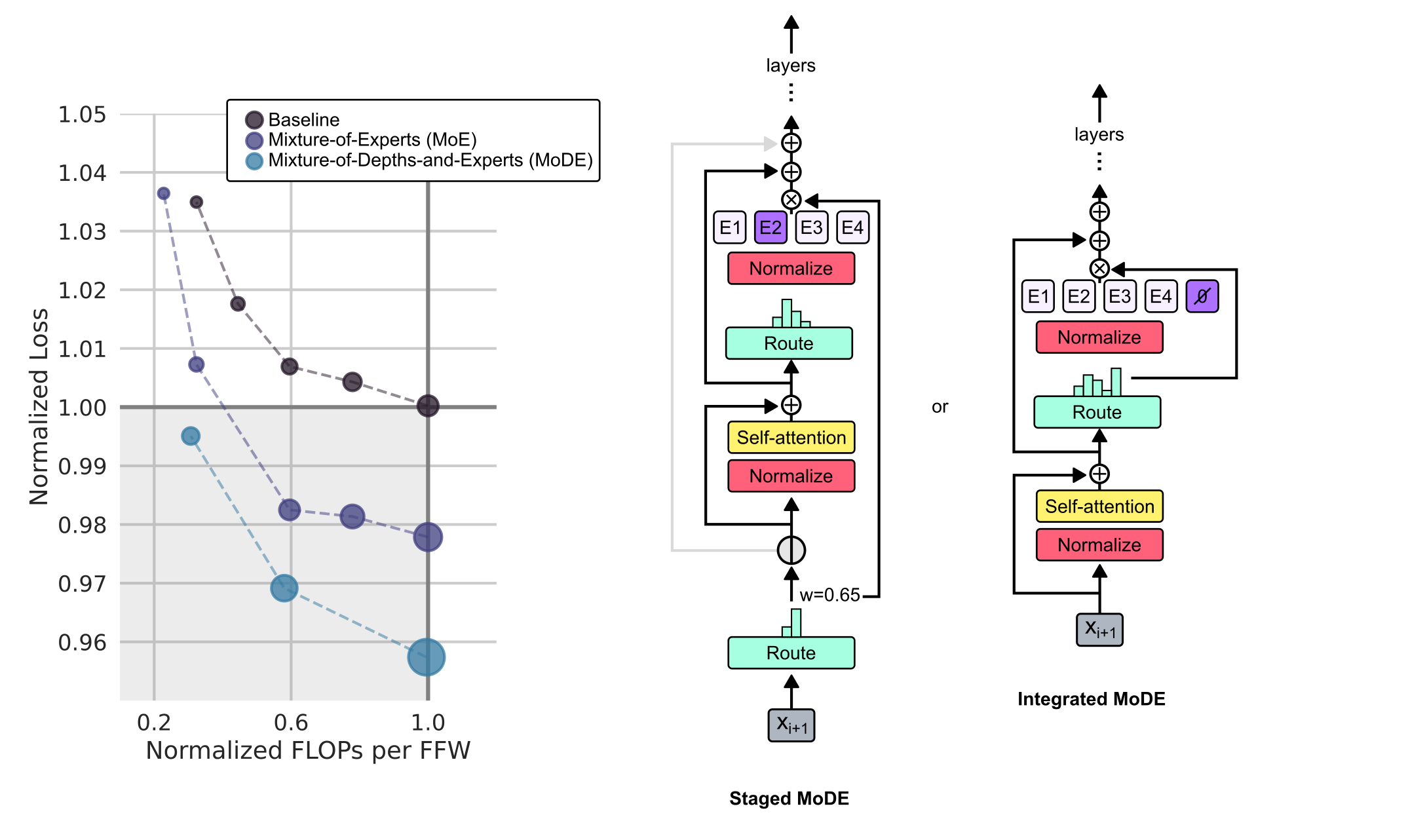}
    \caption{\textbf{Mixture-of-Depths-and-Experts (MoDE).} The MoD technique can be implemented alongside MoE (together comprising MoDE models) in two straightforward manners: staged, which first implements MoD machinery prior to MoE machinery, and integrated, which uses one routing operation to funnel tokens to either experts or no-op operations.}
    \label{fig:mode}
\end{figure}
The MoD technique can be naturally integrated with MoE models (together comprising MoDE models) in addition to vanilla transformers. In figure \ref{fig:mode} we present results showing that the performance improvments offered by MoD compound with those of MoE. We tried two variants: in staged MoDE, which routes tokens around or towards blocks prior to the self-attention step, and integrated MoDE, which implements MoD routing by integrating ``no-op'' experts among the conventional MLP experts. The former is advantageous because it allows for tokens to skip the self-attention step, while the latter is advantageous because it simplifies the routing machinery. We noticed that implementing MoDE in the integrated manner was distinctly better than simply reducing the capacity of experts in conventional MoE models, and relying on token dropping to implement residual routing. We believe this is because with the integrated MoDE machinery, tokens explicitly learn to choose the residual path around the experts, as opposed to preferring an expert but being dropped when implemented as a capacity reduction. 

\section{Discussion}
Mixture-of-Depths transformers empirically demonstrate that one can improve on isoFLOP-optimal baseline performance with models that use fewer FLOPs per forward pass. This means that---for a given training FLOP budget---we can train models that are both faster and better performing than their baseline counterparts. Previously, to train models that are both faster and as- or better-performing than isoFLOP-optimal models, one would have to use surplus compute to \emph{overtrain} smaller models (notably, this overtraining technique is still possible with MoD transformers, and speed gains should compound). 

While MoD transformers require fewer FLOPs per forward pass, one cannot forego FLOPs indiscriminately. Rather, it is crucial to use learned routing decisions---much like in Mixture-of-Experts transformers---to determine whether a token should participate in self-attention and the subsequent MLP (requiring FLOPs), or not (saving FLOPs).We can then use any saved FLOPs by, for example, making the model bigger or training it for longer. Our results show that indeed FLOPs may be inefficiently used in vanilla transformer models, and that there may be more efficient ways for them to be expended.

Learned routing mechanisms are sometimes \textit{non-causal}; that is, information about the future is used to determine a given token's routing decision. This is generally true for top-k routing mechanisms, which are useful because they forego the need for auxiliary balancing losses. However, top-k routing mechanisms present difficulties in post-training autoregressive sampling, where it is impossible to use information about future token identities to determine routing decisions. In this work we show that one can successfully use a top-k routing scheme during training, but not require it during later autoregressive sampling. Eiher a simple auxiliary classifier, or auxiliary loss on the router, is sufficient to learn the top-$k$ routing decisions such that it can mimic the top-$k$ decisions during autoregressive sampling, with minimal to no performance degradation.

Intuitively, a token might learn to route around blocks because the prediction being made at that step is easier, and hence, does not require as much compute. However, this strategy is undoubtedly not all that the network learns. If a token does not participate in self-attention at a certain block, then later tokens will also not be able to attend to \textit{it}. Thus, whether tokens decide to route or not impacts both the current step's prediction and future predictions via causal self-attention, and how the network balances these effects is guided by their influence on the overall language modeling objective. 

This insight opens the door to MoD variants that decouple the routing for queries, keys and values. For example, perhaps a token would prefer to be among the queries, but not the keys, for a given self-attention computation. One can imagine extending this idea even further into the domain of "long-term memory": perhaps there are tokens that would be extremely valuable as keys, regardless of whether it is useful for them to also be among the queries at the step of their occurrence. Learned routing could be a powerful mechanism for deciding which tokens these might be, perhaps funnelling them into a long-term memory buffer that is available during future self-attention. One advantage of such an approach to long-term memory is that tokens decide once, at the moment of "memory encoding", whether they should be retrieved in the future. This is more computationally efficient than performing a full content-based lookup across an entire memory buffer for each step in the future, and could be one step towards drastically increasing the context-length available for making a prediction.

Unlike MoE transformers that route between effectively the same computation (usually MLPs), MoD transformers demonstrate the value of routing among different \textit{types} of computations. In this work the types were either the conventional transformer block, or a null computation (functionally equivalent to multiplying by zero). However, one can imagine extending this idea further by routing between even more types of computation. For example, perhaps some tokens are routed to "memory lookup" functions, and others are routed to "tool use" functions. In general, the routing machinery we deployed provides a knob for adjusting the types of computations available to the network and their relative cost (in total FLOPs); if one wants to introduce an expensive computation, then this can be offset by setting its capacity to some small amount, and hence, by routing only a small number of tokens to it.

Altogether, MoD transformers are another tool one can use to tune a model's compute per forward pass (and hence inference time). The machinery used to implement MoD is also generic, and opens the doors to many extensions and integration with other techniques, such as MoE.
\bibliography{main}

\end{document}